\PassOptionsToPackage{expansion=false}{microtype}
\documentclass[10pt, a4paper]{article}
\usepackage[final]{lrec2026}       % LREC 2026 style with review option

\usepackage[utf8]{inputenc}
\usepackage[T1]{fontenc}
\usepackage{microtype} 

\usepackage{CJKutf8}

\newcommand{\langjp}[1]{%
\begin{CJK*}{UTF8}{min}%
\microtypecontext{activate=false}% This completely shields CJK text from microtype
#1%
\end{CJK*}%
}
\usepackage{booktabs}
\usepackage{tabularx}
\usepackage{multirow}
\usepackage{amsmath}
\usepackage{graphicx}
\usepackage{hyperref}
\usepackage{url}
\usepackage{natbib}
\usepackage{placeins}
\usepackage[table]{xcolor}

\usepackage{newunicodechar}
\newunicodechar{ }{\ } % Safely handles the narrow non-breaking space character

\title{When Irregularity Helps: A Subclass Analysis of Inductive Bias \\in Neural Morphology}
\name{Wen Zhang}
\address{\href{mailto:wenzhang0222@gmail.com}{\texttt{wenzhang0222@gmail.com}}}

\abstract{
Neural morphological generation systems often achieve high aggregate accuracy on benchmark datasets, yet such performance can conceal systematic errors concentrated in rare morphological subclasses. We examine Japanese past-tense verb inflection and show that a very small, structurally specific irregular subtype (<1\% of data) accounts for a disproportionate share of model errors. Controlled ablation experiments demonstrate that removing this subtype yields larger improvements in generalization than removing all irregular verbs, indicating that not all irregularity contributes equally to model instability. These findings suggest that error concentration is driven by the interaction between extreme low-frequency morphological patterns and specific morphophonological processes, particularly gemination. We argue that morphological evaluation should incorporate finer-grained subclass analysis beyond standard conjugation categories.
\\ 
\newline
}

\begin{document}
\maketitleabstract

\section{Introduction}

Neural sequence-to-sequence models have achieved strong performance on morphological inflection benchmarks \citep{kann2016, cotterell2017,wu-etal-2021-applying, vylomova-etal-2020-sigmorphon}. Prior work has emphasized cross-linguistic generalization, low-resource learning, and compositional modeling of unseen lemmas \citep{cotterell2017, kann2016, faruqui-etal-2016-morphological}. However, such evaluations typically report aggregate accuracy, which can obscure systematic weaknesses concentrated in rare morphological subclasses \citep{kirov-etal-2018-unimorph, forster-etal-2021-searching}.

Morphological subclasses encode historically layered and orthographically distinct transformation patterns that differ in frequency and structural complexity. When neural models disproportionately fail on rare subclasses, they exhibit distributional bias toward majority morphological patterns. This is a manifestation of the 'long tail' challenge in morphological learning \citep{cotterell-etal-2018-conll}, where models favor high-frequency regularities over infrequent but productive rules. These effects are analogous to subgroup disparities in general machine learning settings, where performance varies across underrepresented data strata \citep{sagawa2020}, although in this work the “subgroups” are defined purely in terms of linguistic structure rather than social categories.

In this paper, we present a subgroup-aware analysis of Japanese past-tense inflection. We show that a small irregular subclass accounts for a disproportionate share of errors despite representing less than 1\% of training data. Controlled ablation experiments reveal that selective removal of this subtype yields greater improvements in generalization than removing all irregular verbs. These findings indicate that not all irregularity is equally disruptive and that fine-grained subclass analysis is crucial for understanding neural model behavior in morphologically rich languages.

Our contributions are threefold:

\begin{itemize}
\item A structural subgroup evaluation framework for morphological inflection.
\item Quantitative analysis of error concentration across verb classes.
\item Controlled ablation experiments demonstrating selective low-frequency effects and their interaction with model inductive biases.
\end{itemize}

\subsection{Japanese Past-Tense Morphology}

Japanese verbs are traditionally classified by inflectional behavior. Past-tense formation involves the suffix \langjp{た} \textit{-ta} and morphophonological alternations such as consonant mutation, gemination, and vowel changes. These alternations are systematically reflected in \textit{hiragana} orthography, making past-tense inflection a dense testbed for studying morphophonological learning. Because these alternations are encoded transparently in \textit{hiragana}, orthographic variation directly reflects morphophonological structure, allowing character-level models to learn alternation patterns without lexical segmentation. Following \citet{zhang2026mind}, anchoring our inputs in a uniform \textit{kana} space isolates core structural learning and prevents character-to-phoneme mapping errors from confounding the analysis.

\subsection{Canonical Classes}

\textbf{\textit{Godan} (\textit{u}-) verbs} form the past tense through suffix-conditioned stem alternations, often involving consonant changes and gemination. For example, \langjp{かく} \textit{kaku} ‘to write’~$\to$ \langjp{かいた} \textit{kaita} ‘wrote’.  

\noindent\textbf{\textit{Ichidan} (\textit{ru}-) verbs} exhibit stable stems; past tense is formed via direct suffix attachment: \langjp{たべる} \textit{taberu} ‘to eat’~→ \langjp{たべた} \textit{tabeta} ‘ate’, \langjp{みる} \textit{miru} ‘to see’~→ \langjp{みた} \textit{mita} ‘saw’.

\noindent\textbf{Canonical irregular verbs} are a small closed class including \langjp{する} \textit{suru} ‘to do’~→ \langjp{した} \textit{shita} ‘did’~and \langjp{くる} \textit{kuru} ‘to come’~→ \langjp{きた} \textit{kita} ‘came’.

The systematicity of Japanese past-tense morphology makes it an ideal testbed for neural sequence-to-sequence learning, as it provides a controlled environment to probe how models capture both regular inflectional rules and irregular patterns \citep{Yao2018TopicsIN}

\section{Data}

We use a Japanese verb inflection dataset formatted according to SIGMORPHON conventions \cite{vylomova2020,goldman2023}. All forms are converted to \textit{hiragana} to maintain orthographic consistency. Each instance consists of three TAB-separated fields: lemma, target form, and a placeholder indicating that no explicit morphosyntactic features are provided. The model must learn the mapping from lemma to inflected form directly.

\medskip

\noindent Example:

\medskip

\begin{center}
\langjp{ねがえる} \quad \langjp{ねがえった} \quad \_
\end{center}

\medskip

\subsection{Verb Classification}

Verbs are classified according to traditional Japanese conjugation classes, refined to capture orthography-sensitive variation. Canonical irregular verbs (\langjp{する} \textit{suru} and \langjp{くる}\textit{kuru}) and polysemous lemmas with multiple inflected forms are excluded to maintain a clear one-to-one lemma–form mapping. We categorize verbs into four types:

\begin{itemize}
    \item \textbf{Type 1 (\textit{Godan} verbs):} Regular \textit{u}-verbs exhibiting productive, suffix-conditioned stem alternations.  
    Example: \langjp{かく} \textit{kaku} `to write' → \langjp{かいた} \textit{kaita} `wrote'. Count: 2,503

    \item \textbf{Type 2 (\textit{Ichidan} verbs):} Regular \textit{ru}-verbs with stable stems and predictable suffix attachment.  
    Example: \langjp{たべる} \textit{taberu} `to eat' → \langjp{たべた} \textit{tabeta} `ate'. Count: 1,298

    \item \textbf{Type 3 (Canonical irregular verbs):} Excluded in this dataset. Count: 0

    \item \textbf{Type 4 (Other irregular verbs):} Verbs deviating from standard \textit{Godan} or \textit{Ichidan} patterns. Subtypes capture finer-grained orthographic variation:
    \begin{itemize}
        \item \textbf{Type 4-1:} Stem-final /i/ + gemination. These verbs resemble Type 2 but involve gemination at the boundary between the stem-final /i/ and past-tense suffix \textit{-ta}.  
        Example: \langjp{まじる} \textit{majiru} `to mix' → \langjp{まじった} \textit{majitta} `mixed'. Count: 119

        \item \textbf{Type 4-2:} Stem-final /e/ + gemination. Gemination occurs at the boundary between stem-final /e/ and the past-tense suffix \textit{-ta}, increasing structural complexity.  
        Example: \langjp{あきれかえる} \textit{akirekaeru} `to be shocked' → \langjp{あきれかえった} \textit{akirekaetta} `was shocked'. Count: 37

        \item \textbf{Type 4-3:} Localized deviations. These verbs largely follow Type 1 formation but include idiosyncratic stem behavior. Due to the one-to-one lemma–form constraint, only one instance appears.  
        Example: \langjp{いく} \textit{iku} `to go' → \langjp{いった} \textit{itta} `went'. Count: 1
    \end{itemize}
\end{itemize}

\medskip

Table~\ref{tab:dataset} summarizes the dataset distribution across verb types.

\begin{table}[h]
\centering
\small  % smaller font to fit column
\begin{tabular}{lrr}
\toprule
Verb Type & Count & Proportion (\%) \\
\midrule
All verbs & 3,958 & 100 \\
Type 1 (Godan) & 2,503 & 63.2 \\
Type 2 (Ichidan) & 1,298 & 32.8 \\
Type 3 (Canonical irregular) & 0 & 0 \\
Type 4 (Other irregular) & 157 & 4.0 \\
\quad 4-1 (/i/ + gemination) & 119 & 3.0 \\
\quad 4-2 (/e/ + gemination) & 37 & 0.9 \\
\quad 4-3 (localized) & 1 & 0.02 \\
\bottomrule
\end{tabular}
\caption{Dataset statistics by verb type. Subtypes of Type 4 are listed in parentheses for brevity.}
\label{tab:dataset}
\end{table}

\section{Models}

We evaluate two character-level transformer encoder–decoder models for Japanese past-tense inflection. The first follows the SIGMORPHON 2020 baseline \citep{vylomova2020}, and the second is based on the lemma-split evaluation from SIGMORPHON–UniMorph 2023 \citep{goldman2023}, which prevents lemmas from appearing in both training and test sets. Both models operate over \textit{hiragana} strings, capturing consonant gemination, vowel alternation, and other orthographic phenomena, making them suitable for analyzing rare structural irregularities in the dataset.

\section{Experimental Setup}

\subsection{Training Regime}

Training for both models follows the default hyperparameter configurations provided in their respective shared task baselines \citep{vylomova2020, goldman2023}. Models are trained using cross-entropy loss with teacher forcing. Optimization employs the Adam algorithm \citep{kingma2015adam} with standard transformer learning rate scheduling \citep{vaswani2017}. Random seeds are fixed to ensure reproducibility across experiments.

\subsection{Controlled Dataset Conditions}

To systematically evaluate the impact of irregular subtypes on model generalization, we conduct controlled experiments with curated subsets of the dataset:

\begin{itemize}
\item \textbf{Full Dataset (Types 1–4):} Includes all regular and irregular verbs, representing the natural distribution of structural subgroups.
\item \textbf{Regular Only (Types 1–2):} Excludes all irregular verbs, removing orthographically complex minority subgroups.
\item \textbf{Regular + Individual Irregular Subtypes:} Types 1–2 plus one of 4-1, 4-2, or 4-3, isolating the contribution of each irregular subtype.
\item \textbf{Regular + Subtype Combinations:} Selected combinations of 4-1, 4-2, and 4-3 are added to assess interaction effects among minority subgroups.
\end{itemize}

\paragraph{Evaluation Alignment.} For each ablation, the same verb types are removed from training and test sets, ensuring that observed differences reflect the effects of structural subgroups rather than dataset volume.

\subsection{Evaluation Metrics}

Evaluation follows established SIGMORPHON conventions, supplemented by subgroup-level diagnostics to address the limitations of aggregate reporting. Our metrics include:

\begin{itemize}
    \item \textbf{Exact-Match Accuracy:} The percentage of predicted forms identical to gold targets \citep{cotterell2017, goldman2023}.
    
    \item \textbf{Subgroup Accuracy:} Accuracy computed separately for each verb type to reveal concentrated errors in structural minority subgroups, such as rare irregulars, a technique frequently used to diagnose systematic morphological failures \citep{kann2016, makarov2018, vylomova2020}.
    
    \item \textbf{Disparity Ratio:} To quantify the disproportionate concentration of errors across subgroups, we define the \emph{Disparity Ratio} for a subgroup $g$ as:
    \[
    \text{Disparity Ratio}_g = \frac{\text{Error Share}_g}{\text{Data Share}_g}
    \]
    This ratio measures the relative error burden of a subgroup; a value greater than 1 signifies that the subgroup suffers from disproportionately high error rates compared to its prevalence in the data. Similar subgroup-specific performance disparities have been explored in identity-aware AI and general predictive modeling \citep{buolamwini2018, sagawa2020, blodgett2020language}.
\end{itemize}

This evaluation strategy adopts fairness diagnostics from the broader field of NLP, illustrating why aggregate metrics often mask localized weaknesses in linguistic structure \citep{blodgett2020language, lake2018, marcus2018}.

\section{Results}

\subsection{Baseline Performance}

Under full training conditions, both systems achieve high aggregate accuracy on Japanese past-tense inflection:

\begin{itemize}
    \item SIGMORPHON 2020: 97.98\%
    \item SIGMORPHON 2023: 97.73\%
\end{itemize}

Despite high aggregate accuracy, errors are concentrated in specific low-frequency subclasses.

\subsection{Subtype-Specific Ablation Effects}

To assess the contribution of individual irregular subclasses, we remove each Type 4 subgroup independently from both training and evaluation data. Removing Type 4-2 yields the largest performance gains:

\begin{itemize}
    \item 2020: 97.98\% → 99.98\% (+2.00)
    \item 2023: 97.73\% → 99.75\% (+2.02)
\end{itemize}

This corresponds to an error reduction of approximately 99\% relative to baseline error mass for the 2020 system and 88\% for the 2023 system.\footnote{Error reduction computed relative to $(100 - \text{accuracy})$.}

In contrast, removing other irregular subclasses produces substantially smaller improvements. Eliminating all irregular verbs (Type 4) does not produce maximal accuracy, indicating that performance gains are not driven uniformly by irregularity removal.

\subsection{Empirical Distribution of Errors}

\begin{table*}[t]  % * makes it span both columns
\centering
\small  % reduce font size to fit nicely
\setlength{\tabcolsep}{8pt} % adjust column padding
\begin{tabular}{lccc}
\toprule
\textbf{Verb Type} & \textbf{\# Errors 2020} & \textbf{\# Errors 2023} & \textbf{Dominant Error Type} \\
\midrule
1 & 2 & 2 & Gemination / UNK \\
2-1 / 2-2 & 3 & 3 & Gemination \\
4-1 & 1 & 1 & Over-regularization \\
4-2 & 2 & 3 & Gemination \\
\bottomrule
\end{tabular}
\caption{Observed errors under full training conditions for both SIGMORPHON 2020 and 2023 models. Error counts highlight the concentration of failures in low-frequency irregular subtypes.}
\label{tab:error_distribution_merged}
\end{table*}

Table~\ref{tab:error_distribution_merged} summarizes the distribution of observed errors under full training conditions for both SIGMORPHON 2020 and 2023 models. Presenting both models side by side highlights the consistency of error patterns across architectures.

Across both architectures, Type 4-2 verbs account for a disproportionately high share of gemination-related failures relative to their dataset frequency. Most errors involve omission of the small \langjp{っ} \textit{tsu} (e.g., \langjp{あまがけった} \textit{amagaketta} → \langjp{あまがけた} \textit{amagaketa}) or spurious insertion (\langjp{できた} \textit{dekita} → \langjp{できった} \textit{dekitta}). These concentrated failures persist across ablation variants that retain Type 4-2, while removing Type 4-2 sharply reduces total errors. In contrast, Type 4-1 and 4-3 contribute comparatively few errors relative to their dataset share. This asymmetry demonstrates that structural difficulty is driven not by irregular status alone but by the interaction of frequency, phonological conditioning, and orthographic realization.

\subsection{Error Concentration Across Subclasses}

We next examine the distribution of errors by verb type under full training. Although Type 4-2 constitutes only 0.9\% of the dataset, it accounts for 15.8\% of total errors in the 2020 system. 

Disparity Ratios (see Section 4.3.) indicate that Type 4-2 contributes over seventeen times its proportional representation (17.56×) to total errors. By comparison:

\begin{itemize}
    \item Type 1: Ratio 0.80
    \item Type 2: Ratio 0.50
\end{itemize}

Both architectures exhibit consistent error patterns, indicating that concentrated failures in Type 4-2 are model-independent.

Majority subclasses therefore generate fewer errors than expected under a uniform distribution, whereas Type 4-2 generates substantially more. This pattern is consistent in the 2023 system, indicating that it is architecture-independent.

\subsection{Qualitative Error Patterns}

Across both models, the dominant failure pattern in Type 4-2 involves incorrect handling of consonant gemination in past-tense formation. Errors include:

\begin{itemize}
    \item Omission of required gemination (e.g., \langjp{ねがえった} \textit{negaetta} → \langjp{ねがえた} \textit{negaeta})
    \item Spurious gemination (e.g., \langjp{できた} \textit{dekita} → \langjp{できった} \textit{dekitta})
\end{itemize}

These patterns account for the majority of errors within the subclass, suggesting that the instability is structurally localized rather than random.

\section{Error Analysis}

Beyond quantitative accuracy metrics, we conducted fine-grained error analysis across all experimental conditions. We manually examined residual prediction errors from both the 2020 and 2023 models under full and ablated training regimes. Errors were categorized into gemination errors, stem alternation errors, morpheme boundary errors, over-regularization, and vowel length errors. 

As shown in Table~\ref{tab:error_distribution_merged}, Type 4-2 accounts for a disproportionately high share of errors.

\subsection{Error Taxonomy}

Table~\ref{tab:error_taxonomy} presents the error taxonomy, highlighting the orthographic phenomena underlying each error type and their dominant structural sources.

\begin{table*}[t]
\centering
\resizebox{\textwidth}{!}{%
\begin{tabular}{l l l l}
\hline
\textbf{Error Type} & \textbf{Description} & \textbf{Orthographic Property} & \textbf{Dominant Source} \\
\hline
Gemination error & Omission or insertion of small \langjp{っ} & Consonant doubling & Type 4-2 \\
Stem alternation error & Failure to apply expected stem change & Suffix-conditioned alternation & Type 4-2, 4-3 \\
Morpheme boundary error & Misaligned suffix attachment & Boundary detection & Mixed \\
Over-regularization & Irregular treated as regular & Pattern generalization & 4-1, 4-3 \\
Vowel length error & Spurious insertion or deletion & Moraic timing & Rare \\
\hline
\end{tabular}%
}
\caption{Error taxonomy with dominant structural sources and orthographic properties.}
\label{tab:error_taxonomy}
\end{table*}

\subsection{Gemination and Stem Alternation Errors}

Gemination errors are the most frequent and structurally revealing. They arise from omission or spurious insertion of the small \langjp{っ} \textit{tsu} character, particularly in verbs where gemination interacts with stem-final /e/ vowels. 

Under full training, Type 4-2 verbs account for the majority of gemination-related failures despite constituting less than 1\% of the dataset. This concentration is visible in both 2020 and 2023 systems, and persists under multiple ablation regimes unless Type 4-2 itself is removed.

Stem alternation errors occur when the model fails to apply expected vowel-conditioned alternations (e.g., \langjp{あきれかえる} \textit{akirekaeru} → \langjp{あきれかえう} \textit{akirekaeu}). These errors are primarily associated with irregular subclasses 4-2 and 4-3, indicating that alternation and gemination interact in destabilizing ways for the model.

\subsection{Effects of Selective Ablation on Error Patterns}

Ablation results further clarify this pattern. When all irregular verbs are removed, overall error count decreases moderately. However, when only Type 4-2 is removed, error count decreases more substantially, and gemination-related failures nearly disappear.

By contrast, removing 4-1 or 4-3 while retaining 4-2 does not eliminate concentrated gemination errors. This asymmetry indicates that structural complexity is not uniformly distributed across irregular subclasses.

\section{Discussion}

Our analysis demonstrates that irregularity is not uniformly detrimental to neural morphological learning. Instead, its impact depends on structural complexity, distributional frequency, and interaction with the model’s inductive biases.

A specific low-frequency irregular subtype emerges as a structurally distinct case that disproportionately contributes to overall error mass. Despite representing less than 1\% of the training data, it accounts for a substantial share of systematic failures. This concentration suggests that instability arises from specific morphophonological configurations rather than irregularity in general.

Crucially, removing the entire irregular set (Type 4) does not maximize performance. Retaining other irregular subtypes (4-1 and 4-3) produces lower error rates than a purely regular training regime. This suggests a non-monotonic relationship between structural variability and generalization. However, extremely low-frequency, structurally idiosyncratic patterns—such as Type 4-2—are associated with reduced generalization stability.

From a distributional perspective, Type 4-2 functions as a rare but highly influential morphological pattern. Although it represents less than 1\% of the training data, it contributes disproportionately to model errors. Aggregate accuracy (~98\%) obscures this effect, which becomes visible only under subtype-level evaluation.

This pattern highlights a general limitation of aggregate evaluation in morphological modeling: overall performance can mask concentrated weaknesses in rare but structurally complex subclasses. Only fine-grained analysis reveals these effects.

Methodologically, these results suggest that evaluation in morphological generation should incorporate subtype-level reporting and explicit measures of error concentration. Aggregate metrics alone may conceal linguistically meaningful weaknesses, particularly in morphologically rich languages where structural subclasses vary significantly in frequency and complexity.

\section{Conclusion}

We presented a subgroup-aware analysis of Japanese past-tense inflection, examining how minority structural subclasses influence neural generalization. Through controlled ablation experiments, we showed that:

\begin{itemize}
\item Type 4-2 irregular verbs constitute a low-frequency morphological subclass with disproportionate error concentration.
\item Removing only this subtype improves performance more than removing all irregular verbs.
\item Moderate irregularity may support generalization, while extremely low-frequency irregular patterns can destabilize learning.
\end{itemize}

These findings demonstrate that high aggregate accuracy can mask structural effects within morphological systems. Evaluation should therefore go beyond overall performance and include analysis of rare morphological subclasses.

More broadly, these results underscore the importance of distribution-sensitive evaluation in neural NLP. Robust language modeling requires not only high average performance but stable generalization across rare and structurally complex subclasses. Incorporating fine-grained diagnostic analysis into morphological benchmarking can improve both transparency and linguistic validity of evaluation.

\section{Limitations}

Several limitations should be acknowledged.

First, our study focuses on a single language and a single morphological task (past-tense inflection). Although Japanese provides a controlled environment for examining structural effects in morphological learning, cross-linguistic validation is necessary to determine generality.

Second, we evaluate two transformer-based architectures derived from shared tasks. While these represent strong baselines, alternative architectures—such as pretrained character-level language models or multilingual systems—may exhibit different sensitivity to low-frequency morphological patterns.

\section{Future Work}

Several extensions follow naturally from this study.

\paragraph{Cross-linguistic validation.}
Applying the selective-ablation framework to other languages with rich morphology or complex orthographic systems would clarify whether rare morphological subclasses consistently produce disproportionate error concentration across languages.

\paragraph{Architectural comparisons.}
Future experiments should evaluate pretrained character-level models, multilingual encoder--decoder systems, and parameter-efficient adaptation techniques to determine whether error concentration on low-frequency morphological patterns persists under larger-scale pretraining.

\paragraph{Curriculum learning.}
A controlled curriculum that introduces irregular subtypes progressively may illuminate how selective exposure influences generalization. Such experiments could clarify whether rare irregular patterns are inherently difficult or simply poorly learned under uniform training regimes.

\paragraph{Evaluation metrics.}
Developing standardized subgroup-level reporting protocols for morphological generation could promote more transparent evaluation practices across languages and tasks.

Together, these directions aim to improve the evaluation of neural morphological systems by encouraging finer-grained analysis beyond aggregate accuracy. Neural models should be assessed not only by overall performance but also by their robustness across rare and structurally complex morphological patterns.

\section{Acknowledgments}
We thank the reviewers and colleagues for their feedback.

\section{References}
\vspace{-0.7em}  % adjust this value if needed

\bibliographystyle{lrec2026-natbib}
\bibliography{lrec2026-example}
\nocite{*}

\end{document}